\documentclass[letterpaper,10pt,conference]{ieeeconf}

\IEEEoverridecommandlockouts
\usepackage{cite}
\usepackage{amsmath,amssymb}
\usepackage{graphicx}
\usepackage{booktabs}
\usepackage{xcolor}
\usepackage{url}
\usepackage{array}
\usepackage{soul}
\title{\LARGE \bf
A Reconfigurable Dual-Opposition Architecture for Single-Hand Assembly and Manipulation
}

\author{
William Su$^{1}$,
Yunosuke Nakamura$^{2}$,
Yixiao Wang$^{2}$,
Yitong Li$^{3}$,
Mingrui Yu$^{2}$,
Huanan Qi$^{3}$,
Boyuan Liang$^{2}$, \\
Masayoshi Tomizuka$^{2}$,
and Jianshu Zhou$^{3}$%
\thanks{$^{1}$Aerospace Engineering Program, University of California, Berkeley,
Berkeley, CA 94720, USA.}%
\thanks{$^{2}$Department of Mechanical Engineering, University of California, Berkeley,
Berkeley, CA 94720, USA.}%
\thanks{$^{3}$Department of Mechanical Engineering, National University of Singapore,
Singapore 117575.}%
}

\usepackage{booktabs,tabularx,array}

\begin{document}

\raggedbottom
\maketitle
\thispagestyle{empty}
\pagestyle{empty}

\begin{abstract}

In-hand assembly is constrained by the need to maintain grasps on two separate parts while controlling their relative motion within a single hand. To enable both in-hand assembly and manipulation, we present a reconfigurable dual-opposition architecture. Specifically, to support simultaneous grasping of two parts and coordinated in-hand manipulation, four independently actuated fingers are organized into two virtual finger (VF) oppositions, with their relative configuration controlled by a reconfigurable palm. To describe hand motion and simultaneous two-object grasping configurations, a kinematic model of the fingers and palm and an object-size-conditioned workspace formulation are built. To further evaluate motion performance and assembly capability, finger-joint motion and palm tracking are characterized, and in-hand assembly is demonstrated through tasks involving grasping, alignment, fastening, and pressing. Ablation experiments further demonstrate the importance of finger abduction/adduction and palm reconfiguration for successful in-hand assembly. In simulation, the proposed hand achieves a mean continuous sphere rotation success rate of 98.6\% over diameters of 40–230 mm, compared with 73.8\% for the LEAP Hand. After policy fine-tuning with external disturbances, the proposed hand achieves 92.8\% success under disturbances from multiple directions, compared with 45.2\% for the LEAP Hand. Hardware demonstrations further show in-hand rotation of objects of different sizes using policies trained in simulation. Together, these results show that the proposed architecture supports both assembly of two separately held parts and coordinated manipulation of a single object within one hand.

\end{abstract}


\section{Introduction}
\label{sec:introduction}

\begin{figure}[!t]
    \centering
    \includegraphics[width=0.96\columnwidth]{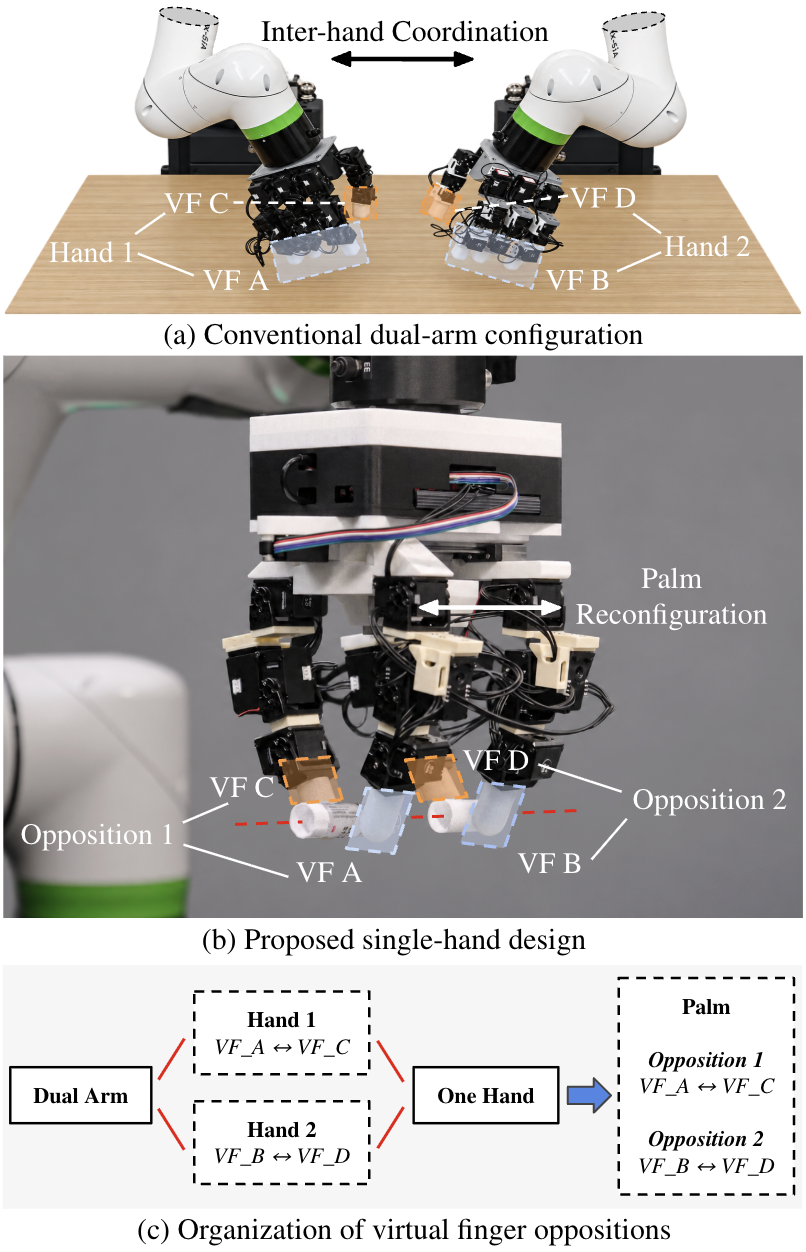}
    \caption{Grasp organization in bimanual and single-hand
configurations. (a) Two anthropomorphic hands form the
oppositions VF A$\leftrightarrow$VF C and
VF B$\leftrightarrow$VF D, respectively.
(b) The proposed hand realizes this grasp organization
with two VF oppositions integrated within one hand.
Palm translation adjusts their relative position.
(c) Schematic mapping from inter-hand coordination to intra-hand reconfiguration.}

    \label{fig:overview}
    \vspace{-2mm}
\end{figure}

\begin{figure*}[t]
  \centering
  \includegraphics[width=\textwidth]{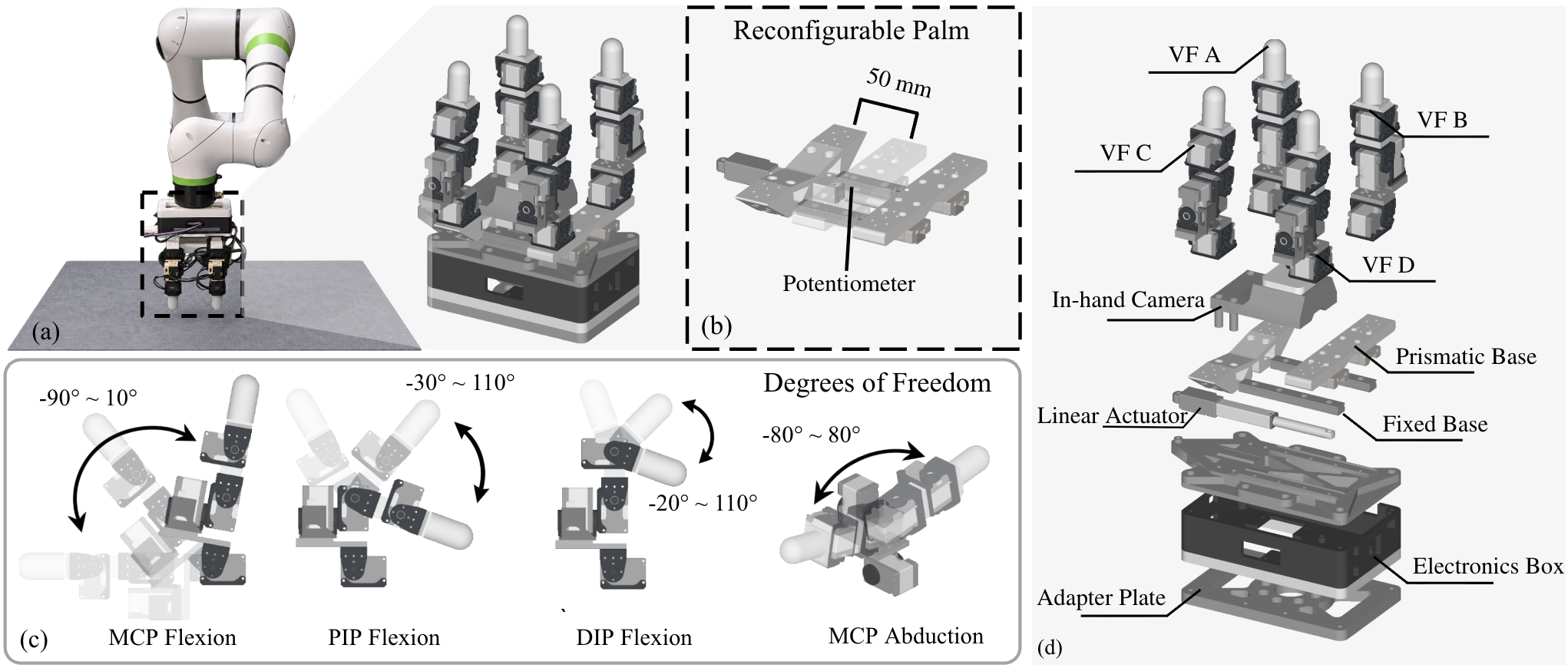}
  \caption{Mechanical design of the proposed 17-DOF reconfigurable hand. (a) The proposed hand mounted on a FANUC CRX-10iA robot arm. (b) Reconfigurable palm mechanism with potentiometer-based position feedback. (c) Finger kinematics showing the four degrees of freedom of each finger, including MCP flexion, PIP flexion, DIP flexion, and MCP abduction/adduction, together with their corresponding motion ranges. (d) Exploded view of the hand.}
  \label{fig:mechanical_design}
\end{figure*}

Dexterous in-hand manipulation and assembly impose different mechanical requirements on a robotic hand. In-hand manipulation coordinates several fingers to change the pose of one object while retaining it within the hand ~\cite{pitz2023tactile,qi2023hora}. Assembly instead requires two parts to remain held while their mating features are aligned and driven through the relative motion required for insertion, fastening, or pressing
~\cite{morgan2021vision,negi2025contactmanifold}. When both parts are held by one end effector, rigid motion of the robot arm moves them together. The
relative motion required for assembly must therefore be generated within the hand. The virtual finger (VF) concept provides a functional abstraction for physical digits acting in unison during grasping and manipulation~\cite{feix2015grasp,iberall1997human}. In a thumb-opposed grasp, the thumb and the opposing
fingers can be represented as two VFs forming one
grasp opposition. This perspective motivates the exploration of hand architectures that can provide multiple functional oppositions and coordinate their relative motion for more complex manipulation and assembly tasks.

Existing hand designs provide different subsets of the required capability.
Multi-DOF finger mechanisms support local fingertip adjustment during
grasping and in-hand manipulation
~\cite{shaw2023leap,si2024deltahands,wu2026dexlink}, while reconfigurable
palms change the relative placement of the fingers around a held object
~\cite{zhou2025dexco,lu2021ruth,richardson2025isyhand}. Hands with symmetric
finger layouts and multi-object grasping methods further show that different
subsets of fingers can form simultaneous grasps on separate objects
~\cite{gao2026detachable,li2023multigrasp}. Suction-enabled fingertips have
also been used to maintain object contact through adhesion during
teleoperated in-hand manipulation~\cite{sun2026sleap}. Assembly requires an
additional capability: controlled motion between the held parts. Bimanual
setups generate this motion by coordinating two arm--hand systems, each
holding one part~\cite{almeida2019relative,tian2025fabrica}. Specialized assembly end effectors generate this motion internally using two orthogonal parallel grippers connected by a prismatic joint, although
local motion at each grasp remains confined to the gripper opening direction
~\cite{triyonoputro2018doublejaw}. However, enabling a single hand to independently adjust two held parts for assembly while retaining the ability for coordinated manipulation of one object remains a
challenging problem.

We present a reconfigurable dual-opposition architecture, realized in a 17-DOF dexterous hand (Fig.~\ref{fig:overview}). The hand combines four independently actuated 4-DOF fingers arranged as two VF oppositions with a reconfigurable palm that controls their relative positions.

The contributions of this paper are as follows:

\begin{itemize}
  \item A reconfigurable dual-opposition architecture combining
two independently controllable VF oppositions with
palm reconfiguration to adjust the relative positions
of held parts and the grasp span for in-hand assembly
and manipulation.

\item Experimental validation through teleoperated in-hand assembly, learned manipulation, and ablation studies, demonstrating the roles of finger abduction/adduction and palm reconfiguration in assembly. Compared with a dexterous anthropomorphic hand baseline, our design achieves sustained rotation over a wider range of object sizes and greater robustness under external disturbances.

\end{itemize}

\section{Mechanical Design}
\label{sec:design}

\subsection{Design Requirements}
\label{subsec:requirements}
Supporting both in-hand assembly and manipulation imposes three mechanical requirements: (1) two independently controllable oppositions with local finger dexterity for simultaneously holding and adjusting two separate parts, including lateral adjustment for alignment; (2) controllable relative reconfiguration between the two oppositions to generate the relative motion required for assembly; and (3) an adjustable grasp span that allows the same fingers to coordinate around single objects of different sizes. These requirements motivate the reconfigurable dual-opposition architecture described below.

\subsection{Dual-Opposition Architecture}
\label{subsec:mechanical_design}

The functional capabilities of conventional parallel grippers, dexterous hands, and the proposed dual-opposition hand are compared in Table~\ref{Table_Architecture}. Conventional parallel grippers and typical dexterous hands generally provide a single functional opposition, which is sufficient for grasping and manipulating one object but does not directly support local adjustment of two held parts. In contrast, the proposed hand integrates two VF oppositions within one end effector. This arrangement preserves the local dexterity of a multi-finger hand while supporting simultaneous grasping and local adjustment of two separate parts. In addition, the reconfigurable palm generates 1-DOF relative translation between the two oppositions, while finger articulation provides local alignment and reorientation for in-hand assembly.

\begin{table}[!ht]
\centering
\caption{Comparison of functional capabilities among conventional parallel grippers, dexterous hands, and the proposed hand.}
\label{Table_Architecture}
\scriptsize
\setlength{\tabcolsep}{3pt}
\renewcommand{\arraystretch}{1.1}
\begin{tabular}{@{}>{\centering\arraybackslash}p{0.29\columnwidth}ccc@{}}
\hline
Capability
& Parallel Gripper
& Dexterous Hand
& Proposed Hand \\ \hline
VF oppositions
& Single
& Typically single
& Dual \\
Local dexterity
& Low
& High
& High \\
Two-object adjustment
& Limited
& Limited
& Supported \\
Relative object motion
& Limited
& Limited
&Supported \\
In-hand assembly
& Limited
& Limited
&Supported \\ \hline
\end{tabular}
\end{table}

\subsubsection{Finger Configuration}
\label{subsubsec:finger_module}

The hand uses four independently actuated fingers
arranged as two functional oppositions (Fig.~\ref{fig:mechanical_design}). Fingers A and C form the VF A $\leftrightarrow$ VF C opposition on the fixed base, while fingers B
and D form the VF B $\leftrightarrow$ VF D opposition on the prismatic base.
The two oppositions can simultaneously hold and locally adjust two separate parts or coordinate around one object in a symmetric finger layout.
Following the LEAP Hand design~\cite{shaw2023leap}, each
finger has four degrees of freedom: metacarpophalangeal
(MCP) flexion, MCP abduction/adduction, proximal
interphalangeal (PIP) flexion, and distal interphalangeal
(DIP) flexion. MCP abduction/adduction provides lateral fingertip adjustment for local alignment within each opposition. All fingers share the same link geometry,
with 16 Dynamixel XL330-M288-T servos actuating the
joints. Custom-molded silicone fingertips have a symmetric geometry and provide local compliance during manipulation.

\subsubsection{Reconfigurable Palm}
\label{subsubsec:palm}

Two linear guides constrain the prismatic base to
translation along the palm axis. An Actuonix
L12-50-100-6-I linear actuator provides a nominal
stroke of $50~\mathrm{mm}$, moving the VF B $\leftrightarrow$ VF D opposition relative to the VF A $\leftrightarrow$ VF C opposition. This translation
adjusts the separation between two held parts and
supplies assembly motion while the fingers control
their alignment and orientation. For in-hand manipulation, the palm adjusts the grasp span to accommodate
a wide range of object sizes.

\subsubsection{System Integration}
\label{subsubsec:integration}

The fixed base and linear guides attach to a common
base plate connected to the robot tool flange. A host
computer commands the finger servos through a U2D2
interface and the palm actuator through an ESP32
controller. Joint encoders and the actuator's
potentiometer provide position feedback. An in-hand
camera supplies an additional view during teleoperation.
The assembled 17-DOF hand weighs $1.227~\mathrm{kg}$.
  \section{Kinematic Modeling}
  \label{sec:modeling}

  \subsection{Hand Kinematics}
  \label{subsec:relative_kinematics}

  Each finger has four joints: MCP flexion, MCP
  abduction/adduction, PIP flexion, and DIP flexion.
  The palm displacement $s_r\in[0,50]~\mathrm{mm}$
  translates fingers B and D relative to A and C.
  Fingertip reference positions in the fixed hand-base
  frame $\{H\}$ are obtained from the following kinematic chains:
  \begin{equation}
    \mathbf p_i =
    \begin{cases}
      \mathbf f_i(\mathbf q_i), & i\in\{A,C\},\\
      \mathbf f_i(\mathbf q_i)+s_r\mathbf e_r,
        & i\in\{B,D\},
    \end{cases}
    \label{eq:hand_forward_kinematics}
  \end{equation}
  where $\mathbf q_i$ contains the four finger-joint
  angles, $\mathbf f_i$ maps these angles to the
  fingertip position at full palm retraction, and
  $\mathbf e_r$ is the unit vector along palm extension.
  With finger joints fixed, palm translation therefore
  shifts the VF B $\leftrightarrow$ VF D opposition relative to VF A $\leftrightarrow$ VF C without changing the relative positions of fingers B and D. This kinematic formulation separates local grasp adjustment through finger articulation from relative translation between the two oppositions provided by the palm.

  \begin{figure}[!t]
    \centering
    \includegraphics[width=\columnwidth]{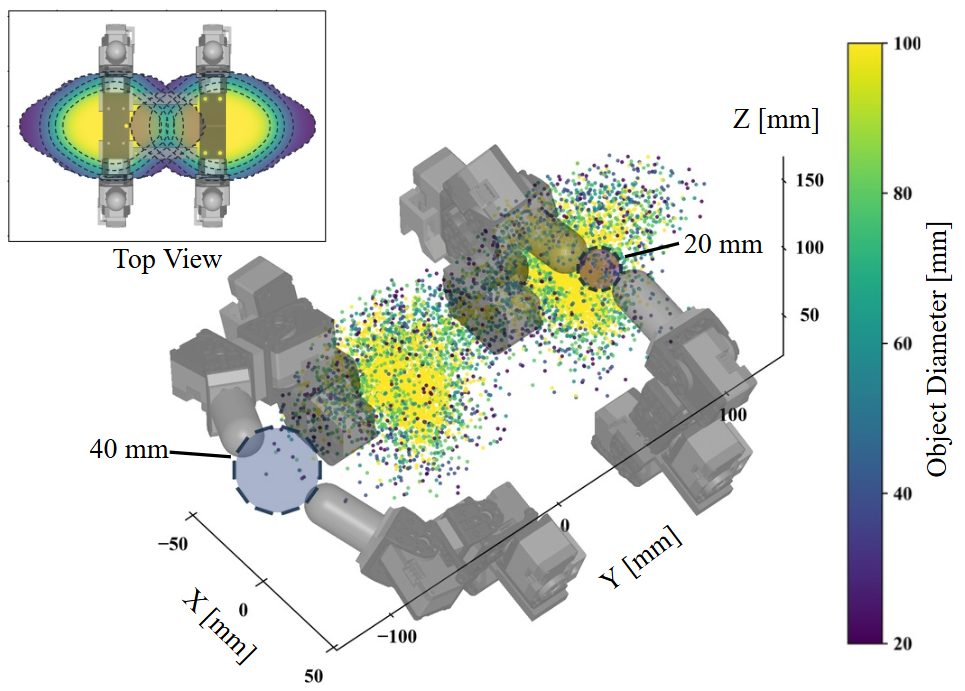}
\caption{Sampled workspace for simultaneous two-object
grasping. Each accepted hand configuration contributes
one sphere center for each VF opposition, with color
indicating sphere diameter. The top view shows
the largest equal sphere diameter accepted in each
spatial bin.}
    \label{fig:dual_object_workspace}
  \end{figure}

  \subsection{Workspace Analysis}
  \label{subsec:contact_workspace}
  \label{subsubsec:object_workspace}

Based on the above kinematic model, we next analyze the feasible grasping workspace of the two oppositions. We use spheres of different diameters to evaluate where
each opposition can hold an object without specifying its orientation~\cite{morrow2021grasping,borras2015workspace,peticco2026karma}.
For each diameter, we identify hand configurations in
which opposing fingertip references are separated by that diameter
within tolerance, and record their midpoint as the
candidate object center. Candidate configurations must satisfy joint and palm
limits and pass collision checks involving the hand
and both objects.

We additionally apply a convex-hull force-closure
test~\cite{lynch2017modern,ferrari1992planning}
to filter the candidate configurations.
Coulomb friction cones are approximated with eight
edges using $\mu=0.6$. A soft-finger model permits
contact moments about the surface normal bounded by
$|\tau_n|\leq\beta f_n$, where $\tau_n$ is the
moment about the contact normal and $f_n$ is the
normal contact force.
We assume $\beta=1~\mathrm{mm}$. Under uniform
pressure over a circular contact patch, the
pure-torsion relation $\beta=2\mu a/3$ gives an
equivalent contact radius of
$a=2.5~\mathrm{mm}$~\cite{vina2016adaptive}.
This radius is a modeling assumption rather than
a measured contact property.
Torque components are normalized by the sphere
radius. Each opposition passes the test when its
primitive wrenches span six dimensions and their
convex hull contains the origin in its strict interior. Figure~\ref{fig:dual_object_workspace} summarizes the
accepted sphere center locations and object sizes.
The three-dimensional point cloud shows the accepted
locations for each opposition, while the top-view
map reports the largest equal sphere diameter
accepted in each spatial bin.
The resulting workspace characterizes feasible
local grasp configurations of the two oppositions
under the assumed contact model. Palm reconfiguration
controls their relative translation along the palm axis.
\section{Experimental Evaluation}
\label{sec:evaluation}

\subsection{Experimental Setup}
\label{subsec:experimental_setup}

In-hand assembly experiments are conducted through
teleoperation. The hand is mounted on a FANUC CRX-10iA
robot arm, and a Meta Quest 3 tracks the operator's
hands. A host computer maps the tracked motion to
finger, palm, and robot arm commands through
geometry-based retargeting. The operator's left
thumb and index finger control the
VF A $\leftrightarrow$ VF C opposition, while the
right thumb and index finger control the
VF B $\leftrightarrow$ VF D opposition.
The distance between the tracked hands controls
the reconfigurable palm. Translation of the
operator's left palm controls robot arm translation.
An in-hand camera provides a secondary view of
regions occluded from the external view.

In-hand manipulation is evaluated in simulation
and on hardware using policies trained through
reinforcement learning. Simulation experiments
compare the proposed hand with the LEAP Hand on
object rotation across different sphere sizes and
under external force disturbances. The policies
trained in simulation are also deployed on the
physical hand to reorient objects of different sizes.
For hardware deployment, the robot arm positions
the hand with its fingers pointing upward, and
the policies control the finger joints and
reconfigurable palm.

\subsection{Hardware Characterization}
\label{subsec:kinematic_performance}

\begin{figure}[!t]
  \centering
  \includegraphics[width=\columnwidth]{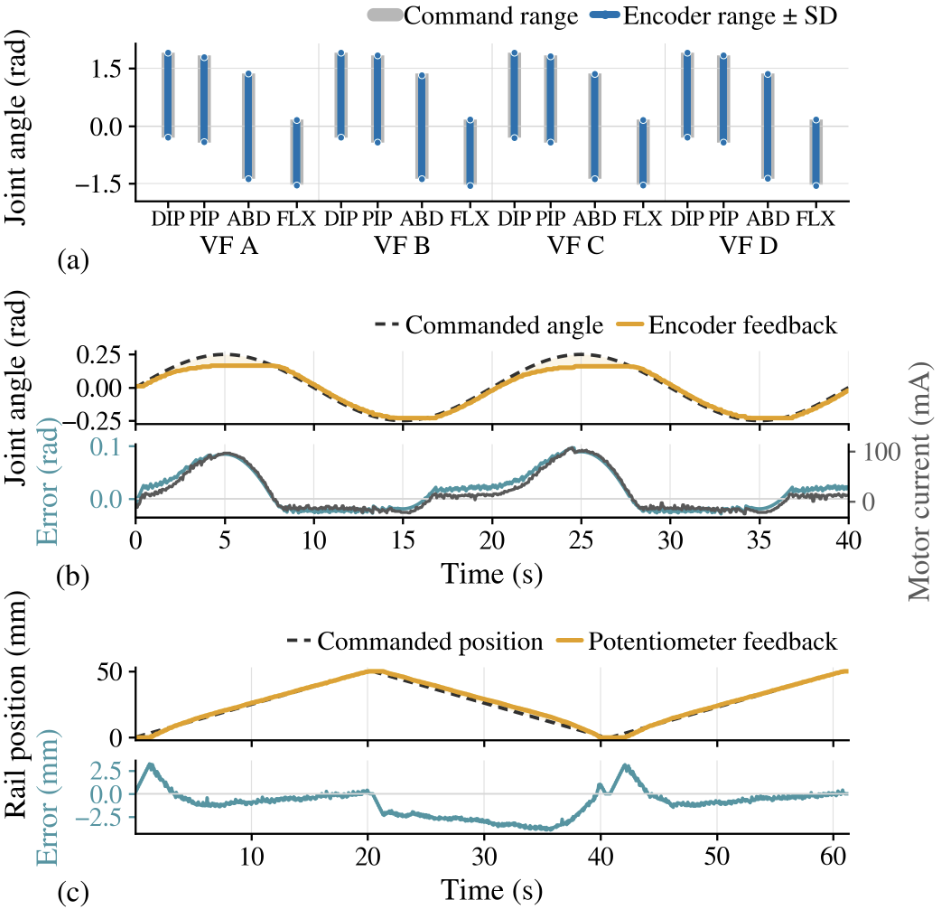}
  \caption{Hardware characterization of the integrated hand.
(a) Commanded joint ranges and encoder measurements;
error bars indicate endpoint standard deviations over
five cycles.
(b) Commanded angle, encoder feedback, tracking error,
and motor current for the MCP abduction/adduction joint
of finger A.
(c) Commanded palm position, potentiometer feedback,
and tracking error during extension and retraction.}
  \label{fig:characterization}
\end{figure}

\begin{figure*}[!t]
  \centering
  \includegraphics[width=\textwidth]{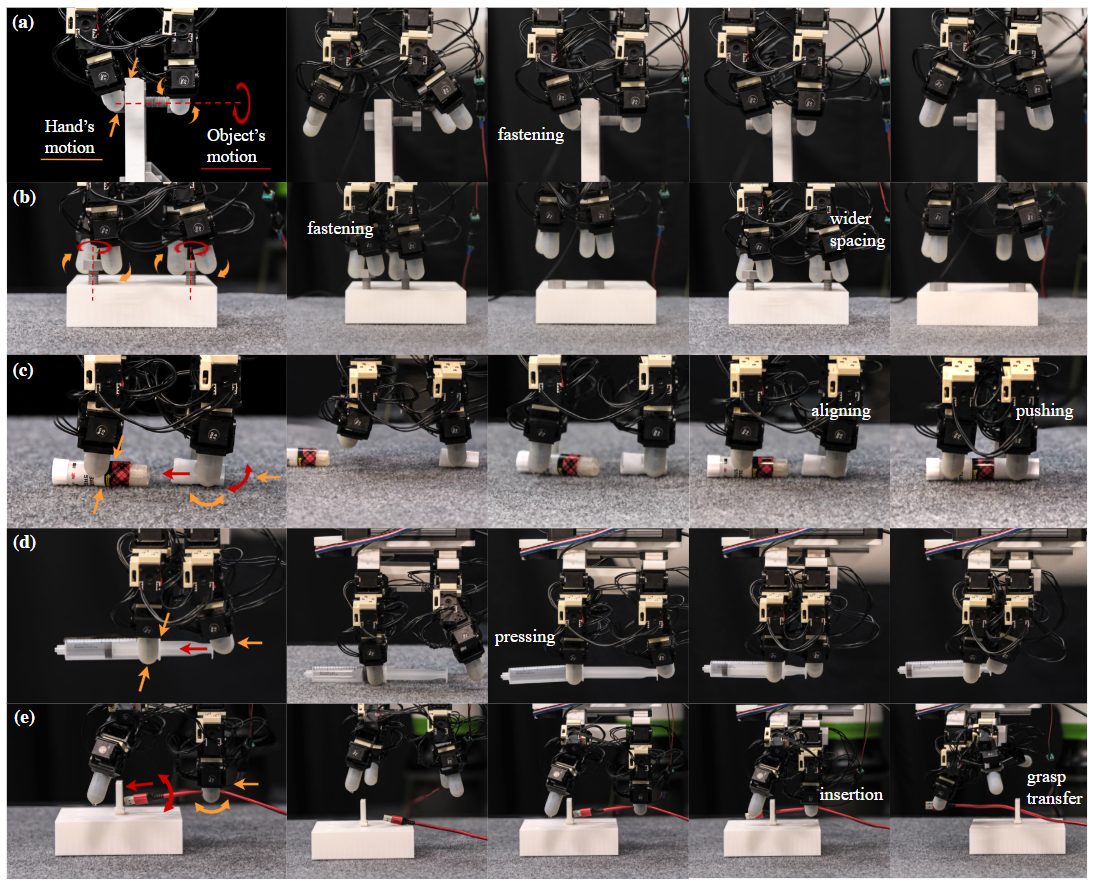}
\caption{Teleoperated assembly demonstrations using the
VF~A~$\leftrightarrow$~VF~C and
VF~B~$\leftrightarrow$~VF~D oppositions.
Each row shows a task sequence progressing from left to right.
(a) Through-wall fastening, with one opposition holding the nut while
the other drives the screw from the opposite side.
(b) Simultaneously fastening two screws at different spacings using
palm reconfiguration.
(c) Aligning and fitting a cap onto a glue-stick body.
(d) Actuating a syringe by holding the barrel with one opposition and
driving the plunger with the other.
(e) Routing a cable through an opening with grasp transfer between the
two virtual-finger oppositions.}
  \label{fig:dual_grasp_tasks}
\end{figure*}

The joint ranges and tracking response of the four fingers
were evaluated after integration into the reconfigurable palm.
With the palm fixed at $s_r=50~\mathrm{mm}$, each of the
16 joints was moved individually through its operating
range shown in Fig.~\ref{fig:mechanical_design} for five
cycles. The range measured by the encoders averaged
$100.14\%$ of the commanded range, and the largest
standard deviation at the endpoints was $0.69^\circ$
(Fig.~\ref{fig:characterization}(a)). Continuous tracking
was evaluated using two cycles of a $0.05$-Hz sinusoidal
command. Figure~\ref{fig:characterization}(b) shows the
MCP abduction/adduction joint of finger A, which had the
largest root mean square error (RMSE) relative to command
amplitude among the 16 joints. At an amplitude of
$0.25~\mathrm{rad}$, its RMSE was $0.041~\mathrm{rad}$,
the ratio of measured to commanded amplitude was $0.848$,
and the response lagged the command by $0.374~\mathrm{s}$.

Palm tracking was evaluated over a continuous
$0\rightarrow50\rightarrow0\rightarrow50~\mathrm{mm}$
trajectory at approximately $2.49~\mathrm{mm\,s^{-1}}$.
Position was measured using potentiometer feedback calibrated
at full extension and retraction. The overall tracking RMSE
was $1.76~\mathrm{mm}$, with a maximum absolute error of
$3.90~\mathrm{mm}$ (Fig.~\ref{fig:characterization}(c)).
The two extension movements produced RMSE values of
$1.00$ and $0.99~\mathrm{mm}$, whereas retraction produced
a larger RMSE of $2.74~\mathrm{mm}$. During the holds
at the endpoints, the maximum difference between the
commanded position and feedback was $0.14~\mathrm{mm}$,
compared with the larger tracking errors during translation. The operating ranges define the joint limits and target
position bounds used during policy training in simulation.
The finger and palm actuator models are calibrated by
matching simulated position trajectories to the recorded
hardware responses under the same commands.

\subsection{In-Hand Assembly via Teleoperation}
\label{subsec:dual_grasp_experiments}

To evaluate how the dual-opposition architecture
supports in-hand assembly, we conduct five
teleoperated tasks involving independent part
adjustment, coordinated holding and pushing,
and grasp transfer
(Fig.~\ref{fig:dual_grasp_tasks}).
The tasks comprise through-wall fastening,
simultaneous screw fastening at different spacings,
glue-stick capping, syringe actuation, and cable
routing through an opening. Together, they examine
how independent control of the two VF oppositions
combines with palm reconfiguration to accommodate
different part spacings and generate relative
motion during assembly.

\textit{1) Through-wall fastening:}
The proposed hand holds a nut on one side of a wall and drives
a screw from the opposite side
(Fig.~\ref{fig:dual_grasp_tasks}(a)).
The VF~A~$\leftrightarrow$~VF~C opposition holds the nut in place,
while the VF~B~$\leftrightarrow$~VF~D opposition rotates the screw.
Translation of the prismatic base follows the screw's axial
advance as it threads into the nut.  This task demonstrates coordinated finger rotation
and palm translation to complete fastening while
the robot arm remains stationary.

\textit{2) Simultaneous screw fastening:}
The two VF oppositions simultaneously fasten
separate vertical screws
(Fig.~\ref{fig:dual_grasp_tasks}(b)).
Before each fastening sequence, palm reconfiguration
sets the separation between the oppositions to match
the screw spacing. Each opposition then rotates its
screw while the robot arm follows their axial
advance, with palm spacing held fixed. This task uses palm reconfiguration to adapt the working span to the screw spacing, whereas
through-wall fastening uses palm translation
to generate the required relative motion during assembly.

\textit{3) Glue-stick capping:}
Figure~\ref{fig:dual_grasp_tasks}(c) demonstrates glue-stick
capping by grasping two separately placed parts and aligning
them for assembly within the hand. The cap and body are
initially placed at different locations on the table.
The VF~B~$\leftrightarrow$~VF~D opposition first grasps the cap,
after which the robot arm positions the
VF~A~$\leftrightarrow$~VF~C opposition to grasp the body.
With both parts held, finger abduction adjusts their relative
alignment. Translation of the prismatic base then presses
the cap onto the body while the robot arm remains stationary.
This task combines local alignment with relative palm translation to bring the held parts together for assembly.

\textit{4) Syringe actuation:}
Figure~\ref{fig:dual_grasp_tasks}(d) demonstrates linear actuation
while one VF opposition holds the barrel.
The VF~A~$\leftrightarrow$~VF~C opposition grasps the syringe barrel,
while fingers B and D close together behind the
plunger to form a pushing contact. Translation of the prismatic base
then drives the plunger into the barrel while the barrel remains fixed. This task demonstrates complementary holding and
pushing actions with one opposition stabilizing the
syringe barrel, while palm translation supplies a pushing force to advance the plunger.

\textit{5) Cable routing:} Fig.~\ref{fig:dual_grasp_tasks}(e) demonstrates routing a cable through an opening
with grasp transfer between the VF~A~$\leftrightarrow$~VF~C
and VF~B~$\leftrightarrow$~VF~D oppositions. One opposition
initially grasps the cable and aligns its end with the opening.
After the cable end is advanced through the opening, the other
opposition grasps the emerging section on the opposite side.
The initial opposition then releases the cable, allowing the
receiving opposition to pull it farther through the opening. This task demonstrates coordinated grasp transfer,
allowing the hand to switch from pushing the cable
through the opening to pulling it from the other side.

\begin{table}[t]
\centering
\caption{Assembly task outcomes for the full hand and
two ablated configurations.
$\checkmark$: success; $\times$: failure.}

\label{tab:assembly_ablation}
\footnotesize
\renewcommand{\arraystretch}{1.10}
\setlength{\tabcolsep}{2.5pt}
\begin{tabular}{@{}lccc@{}}
\toprule
Task
& \shortstack{Full\\hand}
& \shortstack{MCP abduction/\\adduction locked}
& \shortstack{Fixed\\palm} \\
\midrule
Through-wall fastening       & $\checkmark$ & $\checkmark$     & $\times$ \\
Simultaneous screw fastening &$\checkmark$ & $\times$     & $\times$     \\
Glue-stick capping           & $\checkmark$ & $\times$     & $\times$     \\
Syringe actuation            & $\checkmark$ & $\checkmark$ & $\times$     \\
Cable routing              & $\checkmark$ & $\checkmark$ & $\checkmark$ \\
\bottomrule
\end{tabular}
\end{table}

\subsection{In-Hand Assembly Ablation Study}
\label{Ablation}

To evaluate the contributions of MCP abduction/adduction
and palm reconfiguration, we qualitatively compare
the full hand with two ablated configurations:
MCP abduction/adduction locked for all four fingers, or the palm fixed at
$s_r=50~\mathrm{mm}$. All remaining finger joints
and robot arm motions remain available for
teleoperation. Table~\ref{tab:assembly_ablation}
summarizes the task outcomes.

Locking MCP abduction/adduction restricts the hand's ability to make local grasp adjustments. In the tested configurations, finger abduction/adduction is needed to sustain screw rotation during simultaneous screw fastening and to align the cap with the glue-stick body. Removing this motion therefore prevents completion of these two tasks. Tasks involving through-wall fastening and syringe actuation remain successful using the available finger motion and palm translation. Cable routing also remains successful, with robot arm motion compensating for cable alignment and insertion. These outcomes indicate that MCP abduction/adduction is critical for assembly tasks that require dexterous local adjustments of held parts.

Fixing the palm at $50~\mathrm{mm}$ removes the hand's ability to adjust the separation between the two oppositions. During through-wall fastening, the fixed palm cannot follow the screw's axial advance, requiring the
fingers to compensate through abduction/adduction.
The resulting changes in grasp posture make sustained
screw rotation difficult and prevent task completion. The hand also cannot accommodate changes in screw
spacing during simultaneous fastening. Tasks involving glue-stick capping and syringe actuation require relative axial motion while maintaining alignment between the held parts. Attempts to produce this motion through finger abduction/adduction instead alter part alignment and lead to loss of grasp. Cable routing remains successful because robot arm motion compensates for the fixed palm by repositioning the hand for cable insertion and the subsequent grasp transfer.

\begin{figure}[t]
  \centering
  \includegraphics[width=\columnwidth]{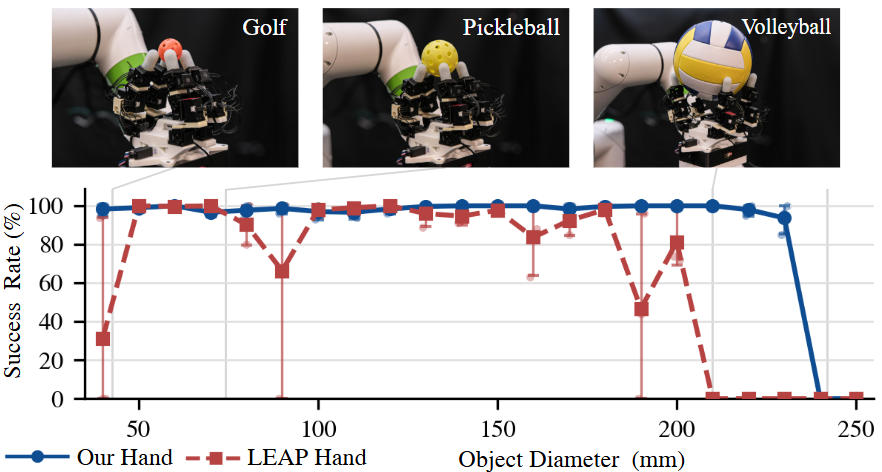}
  \caption{In-hand object rotation in simulation and on hardware.
  Photographs show hardware experiments with a practice golf ball,
  a pickleball, and a volleyball. Large markers show
mean simulation success rates across three training
seeds; small markers show results for each seed.
Error bars indicate 95\% hierarchical bootstrap
confidence intervals.}
  \label{fig:simulation_results}
\end{figure}

\begin{figure*}[!t]
  \centering
  \includegraphics[width=0.98\textwidth]{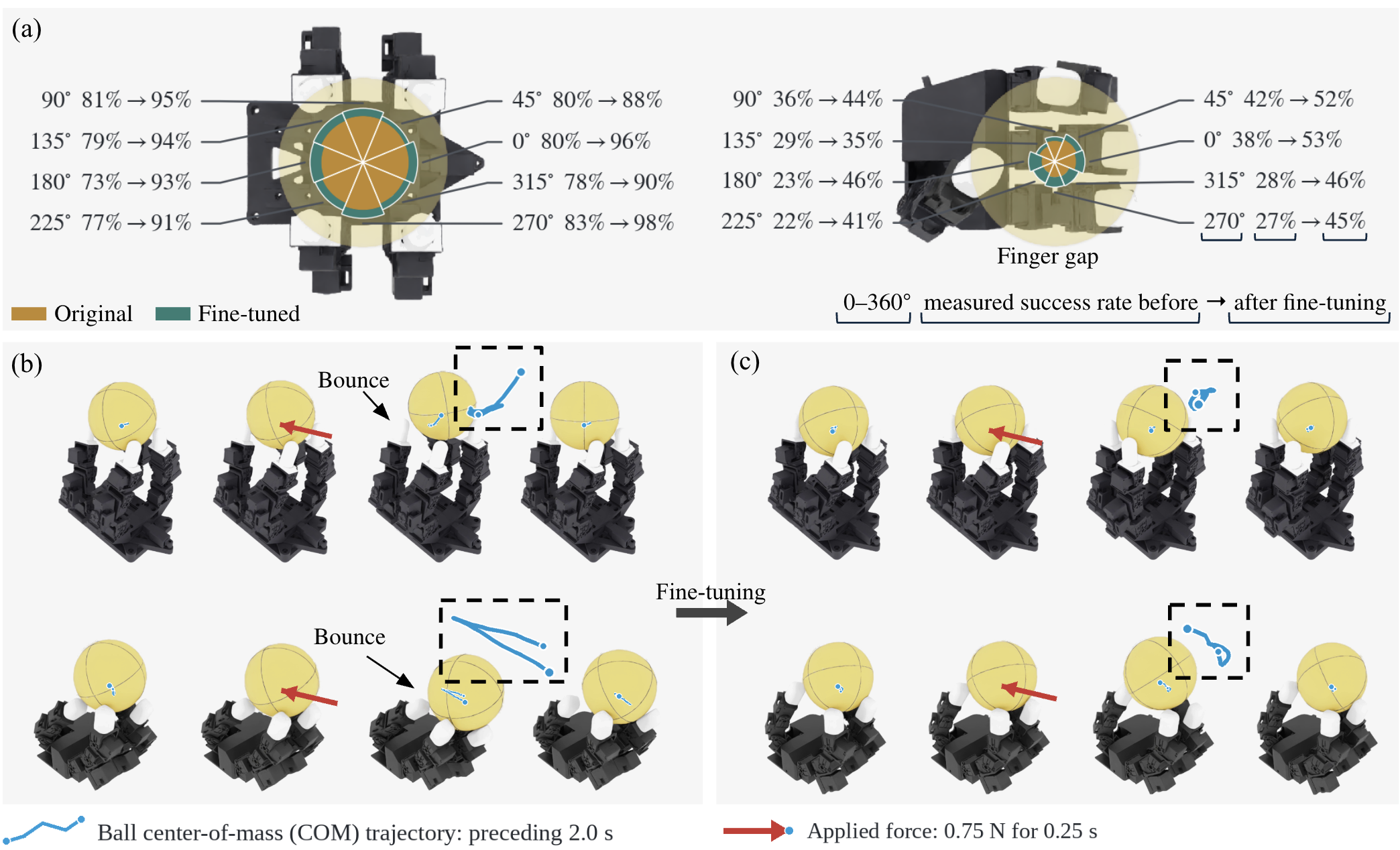}
  \caption{Response to an applied force with a peak magnitude
of $0.75~\mathrm{N}$ and a duration of $0.25~\mathrm{s}$
during rotation of a $100~\mathrm{mm}$ sphere.
(a) Success rate for continuous sphere rotation before and after
fine-tuning under disturbances from eight directions,
with 32 rollouts per direction per seed across three
training seeds of the original policies.
(b), (c) Keyframes from a selected rollout at
$225^\circ$ using the original and fine-tuned policies,
respectively. The top and bottom rows show the proposed
hand and the LEAP Hand, respectively. Frames show the
motion before, during, and after the disturbance.
Blue traces show the sphere COM trajectory over the
preceding $2.0~\mathrm{s}$; red arrows indicate the
applied force.}
  \label{fig:disturbance_results}
\end{figure*}

\subsection{In-Hand Manipulation via Reinforcement Learning}
\label{subsec:four_finger_manipulation}

The proposed hand combines a symmetric arrangement
of two VF oppositions with an adjustable grasp span.
We evaluate its in-hand manipulation capability
through two complementary studies.
The first examines whether rotation can be sustained
across different object sizes ranging from $40$ to $250~\mathrm{mm}$, requiring the fingers
to establish and coordinate contact over different
grasp spans. The second examines whether rotation
can continue after an applied external disturbance.
Both studies compare the proposed hand with the
LEAP Hand in simulation using policies trained
through reinforcement learning. Finally, policies trained in simulation are deployed on the physical
hand to evaluate sustained rotation of
physical objects of different sizes.

\subsubsection{In-Hand Manipulation Across Object Sizes}
\label{subsubsec:object_size_evaluation}
The proposed hand is compared with the right LEAP Hand using 22 thin
spherical shells with diameters from $40$ to $250~\mathrm{mm}$ in
$10~\mathrm{mm}$ increments.
The same object models are used for both hands.
Sphere mass scales with surface area, taking the $210~\mathrm{mm}$,
$0.270~\mathrm{kg}$ volleyball as the reference:
\begin{equation}
  m(d)=0.270\left(\frac{d}{210~\mathrm{mm}}\right)^2,\quad
  I(d)=\frac{2}{3}m(d)\left(\frac{d}{2}\right)^2
  \label{eq:ball_scaling}
\end{equation}

A separate policy is trained for each hand using proximal policy optimization (PPO), with 2048 parallel
environments and 2000 training iterations. Both policies are implemented
as a multilayer perceptron (MLP) with three hidden layers containing
512, 256, and 128 units, respectively. The inputs contain
three consecutive frames of joint positions and command targets,
together with the sphere diameter. The proposed-hand policy controls
the 16 finger joints and the reconfigurable palm, whereas the LEAP
Hand policy controls its 16 finger joints. Following HORA's clipped
angular-velocity objective~\cite{qi2023hora}, the reward encourages
rotation about the commanded axis. Additional penalties discourage
object translation, off-axis rotation, and deviation from the initial
joint configuration, with a separate penalty for dropping the object.
This reward and the same domain randomization settings for contact
friction and finger actuator parameters are used for both hands. Each hand is evaluated across three random training
seeds, with 100 held-out 30-s rollouts per seed and
diameter. A rollout is successful if the hand retains the
sphere and sustains rotation throughout the
30-s evaluation. To assess
whether rotation is sustained, we divide each rollout
into six consecutive 5-s intervals. The mean signed
angular velocity must be positive in every interval,
and the lowest interval mean must be at least 50\%
of the mean over the full rollout.

Across sphere diameters of $40$--$230~\mathrm{mm}$,
the proposed hand achieves a mean success rate of
$98.6\%$, compared with $73.8\%$ for the LEAP Hand
(Fig.~\ref{fig:simulation_results}).
At diameters of $210$, $220$, and $230~\mathrm{mm}$,
the proposed hand achieves success rates of
$100\%$, $98.0\%$, and $93.7\%$, respectively,
whereas the LEAP Hand has no successful rollouts.
These results show that the proposed hand sustains
rotation over a wider range of tested sphere sizes. We further deploy the policies trained in simulation
on the physical hand for in-hand rotation of a
practice golf ball, a pickleball, and a volleyball
(Fig.~\ref{fig:simulation_results}), using the
corresponding object diameter as a policy input.
Each object is manually placed in the initial grasp
before policy execution. The robot arm remains
stationary while the policy controls the finger
joints and reconfigurable palm. The hand maintains
rotation throughout a $20$-s rollout for each object,
demonstrating in-hand manipulation across a wide range of
object sizes on hardware.

\subsubsection{In-Hand Manipulation Under External Disturbances}
\label{subsubsec:disturbance_evaluation}
To compare the ability of the proposed hand and the
LEAP Hand to sustain in-hand manipulation under external
disturbances, we apply an external force to a
$100~\mathrm{mm}$ sphere during rotation. Without
disturbances, the proposed hand and the LEAP Hand
achieve success rates of $97.0\%$ and $98.0\%$,
respectively, across the three training seeds of the original policies.
During each rollout, the force is applied once at
the sphere center of mass (COM) for
$0.25~\mathrm{s}$, with a peak magnitude of
$0.75~\mathrm{N}$ and an onset sampled between
$8$ and $20~\mathrm{s}$. The force acts
$15^\circ$ above the horizontal plane at one of
eight azimuths spaced by $45^\circ$ in the world
frame, with $0^\circ$ aligned with the positive
$x$-axis. We compare the original policies with policies
fine-tuned for 500 iterations. During fine-tuning,
half of the episodes are disturbance-free; the
remaining episodes contain one to four force
applications with a peak magnitude of
$0.5~\mathrm{N}$, random onset times, and azimuths
sampled uniformly in the horizontal plane.
Force observations are not provided to the policies.
Each condition is evaluated using 32 held-out
30-s rollouts per direction per seed across three
training seeds of the original policies. The success criterion remains the same.

Fig.~\ref{fig:disturbance_results}(a) shows a gap
between the thumb and index finger in the illustrated
LEAP Hand grasp, while the proposed hand distributes
its four fingers more evenly around the object.
The original LEAP Hand policies achieve their lowest
success rate of $21.9\%$ at $225^\circ$, which
points toward this gap. In comparison, the proposed
hand achieves $72.9$--$83.3\%$ success across all
eight directions. This more uniform performance
is consistent with the balanced support offered
by its symmetric finger layout. Fine-tuning improves success in all eight directions
for both hands. Overall success increases from
$79.0\%$ to $92.8\%$ for the proposed hand and from
$30.6\%$ to $45.2\%$ for the LEAP Hand. For the
LEAP Hand, gains at the initially weaker directions
of $180^\circ$, $225^\circ$, and $270^\circ$
exceed those at $0^\circ$ and $45^\circ$,
suggesting a more balanced response after
fine-tuning. Success at $225^\circ$ increases
from $21.9\%$ to $40.6\%$. The proposed hand
maintains higher success in all eight directions
before and after fine-tuning.

The keyframes in
Fig.~\ref{fig:disturbance_results}(b,c) show changes in hand
posture and object motion in a selected rollout.
After fine-tuning, the LEAP Hand positions its thumb
higher around the sphere, while the proposed hand
holds the sphere lower relative to the hand plane.
Both hands show smaller peak sphere COM
displacements. The original LEAP Hand policy retains
the sphere but fails to sustain rotation after
the disturbance, whereas the fine-tuned policy
maintains continuous rotation.

\section{Conclusion and Future Work}
\label{sec:conclusion}

A reconfigurable dual-opposition architecture has been developed and realized in a 17-DOF hand for in-hand assembly and manipulation. Teleoperated assembly demonstrations and ablation studies show how finger articulation and palm reconfiguration provide complementary capabilities for local part adjustment and relative motion between held parts. In simulation, the proposed hand sustains continuous object rotation over a wider range of object sizes and achieves greater robustness to disturbances from multiple directions than the LEAP Hand. Hardware experiments further demonstrate in-hand rotation of objects of different sizes using policies trained in simulation. Together, these results demonstrate that the proposed dual-opposition architecture provides a common mechanical basis for both two-part in-hand assembly and coordinated single-object manipulation.

In future work, autonomous in-hand assembly will be investigated through imitation learning from teleoperated demonstrations combined with in-hand visual feedback. We will also systematically investigate how hand morphology and the number and organization of VF oppositions affect in-hand assembly capability, including architectures with more than two oppositions and comparisons with conventional five-finger dexterous hands. Manipulation will be extended to non-spherical objects, and fingertip tactile sensing will be investigated for contact detection, slip detection, and in-hand object pose estimation.

\bibliographystyle{IEEEtran}
\bibliography{references}

\end{document}